\pdfoutput=1
\documentclass[letterpaper,10pt,journal]{IEEEtran}
\usepackage[T1]{fontenc}
\usepackage{amsmath}
\usepackage{amssymb}
\usepackage{booktabs}
\usepackage{array}
\usepackage{longtable}
\newcolumntype{L}[1]{>{\raggedright\arraybackslash}p{#1}}
\usepackage{graphicx}
\usepackage[table]{xcolor}
\definecolor{ourrow}{RGB}{232,241,251}
\usepackage{cite}
\usepackage{url}
\usepackage{algorithm}
\usepackage{algpseudocode}
\usepackage[activate={true,nocompatibility}]{microtype}
\AtBeginDocument{\microtypesetup{activate=false}}
\usepackage[hidelinks]{hyperref}
\newenvironment{IEEEpractitioners}
  {\renewcommand{\IEEEkeywordsname}{Note to Practitioners}\begin{IEEEkeywords}}
  {\end{IEEEkeywords}}

\title{ReContact: Reconstruct and Retarget Human Contacts from Video for Contact-Rich Humanoid Manipulation}
\author{Zhongkai~Gu, Yang~Liu, Wei~Zhu$^{*}$, and Mitsuhiro~Hayashibe$^{*}$%
\thanks{The authors are with the Department of Robotics, Graduate School of Engineering, Tohoku University, Sendai 980-8579, Japan.}%
\thanks{$^{*}$Corresponding authors: Wei Zhu (e-mail: zhu.wei.c1@tohoku.ac.jp) and Mitsuhiro Hayashibe (e-mail: mitsuhiro.hayashibe.e6@tohoku.ac.jp).}}
\hypersetup{pdftitle={ReContact: Reconstruct and Retarget Human Contacts from Video for Contact-Rich Humanoid Manipulation},pdfauthor={Zhongkai Gu, Yang Liu, Wei Zhu, Mitsuhiro Hayashibe},pdfkeywords={Contact-rich manipulation, humanoid robots, learning from demonstration, loco-manipulation, motion retargeting, residual reinforcement learning}}
\newcommand{\TtwoSuccessMean}{83.5}
\newcommand{\TtwoSuccessSD}{5.2}
\newcommand{\TtwoBestComparatorMean}{0.16}
\newcommand{\TtwoCwsBottomLeft}{0.6}
\newcommand{\TtwoCwsBottomRight}{65.7}
\newcommand{\TtwoOursBottomLeft}{80.8}
\newcommand{\TtwoOursBottomRight}{81.2}
\newcommand{\TtwoOursThigh}{94.4}
\newcommand{\TtwoOursNormal}{16.3}
\newcommand{\TtwoOursWork}{544.9}
\newcommand{\TtwoComparatorCompletionMin}{24.3}
\newcommand{\TtwoComparatorCompletionMax}{65.7}
\newcommand{\TtwoComparatorThighMin}{0.2}
\newcommand{\TtwoComparatorThighMax}{36.6}
\newcommand{\TtwoComparatorNormalMin}{47.8}
\newcommand{\TtwoComparatorNormalMax}{55.0}
\newcommand{\TtwoComparatorWorkMin}{565.2}
\newcommand{\TtwoComparatorWorkMax}{868.0}

\begin{document}
\maketitle

\begin{abstract}

Contact transitions allow humanoid robots to redistribute object support across the
whole body, enabling manipulation beyond hand-only grasps. Human videos provide a scalable
source of such transitions for robots to learn from. However, body occlusion disrupts object
pose tracking and corrupts retargeted motion. Moreover, accurately reproducing the object
trajectory does not ensure that the robot applies stabilizing forces at the intended contacts,
risking slips or drops in later phases.
To address these challenges, we propose ReContact, which reconstructs and retargets
human-object interactions through an object-centric contact plan. The plan specifies contact
phases, contact pairs, anchors, pressing directions, and sticking or sliding modes. Contact
priors stabilize object pose reconstruction through occlusion. A residual policy over a frozen
whole-body tracker is rewarded for making contact at planned anchors with forces aligned to
the specified directions.
In simulation, on a thigh-supported regrasp task, ReContact achieves the planned two-palm bottom
hold in $\TtwoSuccessMean\pm\TtwoSuccessSD\%$ of episodes over three training seeds, versus $\TtwoBestComparatorMean\%$
for the best adapted baseline. One policy trained on $247$ references re-solved for new
placements raises in-workspace tube-transfer success from $20.4\%$ (source reference only) to
$97.8\%$. On a real Unitree G1, policies execute seven contact-rich manipulation and
two loco-manipulation tasks. Project page: \url{https://recontact-robot.github.io}.

\end{abstract}

\begin{IEEEpractitioners}
A box picked up by its sides is rested on a thigh and then carried from
below; a tube is tucked under the arm to free a hand. Humanoids need the same skills
to handle goods larger or heavier than their hands can hold. ReContact
turns one human video of such a task
into a contact plan that states which robot surface presses which
object surface, in which direction, whether it stays fixed or slides,
and when support changes; GPT-6
Astra drafts and reviews it, and an operator can correct it in the same
viewer. The plan shapes the whole-body motion, defines the training
rewards and generates references for other object placements, so one
tube-transfer reference yields a policy for a tabletop workspace. Deployment needs
the object's dimensions and pose, a pretrained whole-body tracker and one
trained policy per task family. On a Unitree
G1, the learned skills are chained with walking from table pickup to floor placement; quantitative comparisons are made in simulation.
\end{IEEEpractitioners}

\begin{IEEEkeywords}
Humanoid robots, contact-rich manipulation, learning from demonstration, residual reinforcement learning.
\end{IEEEkeywords}

\section{Introduction}
\label{sec:intro}

\begin{figure*}[!t]
\centering
\includegraphics[width=\textwidth]{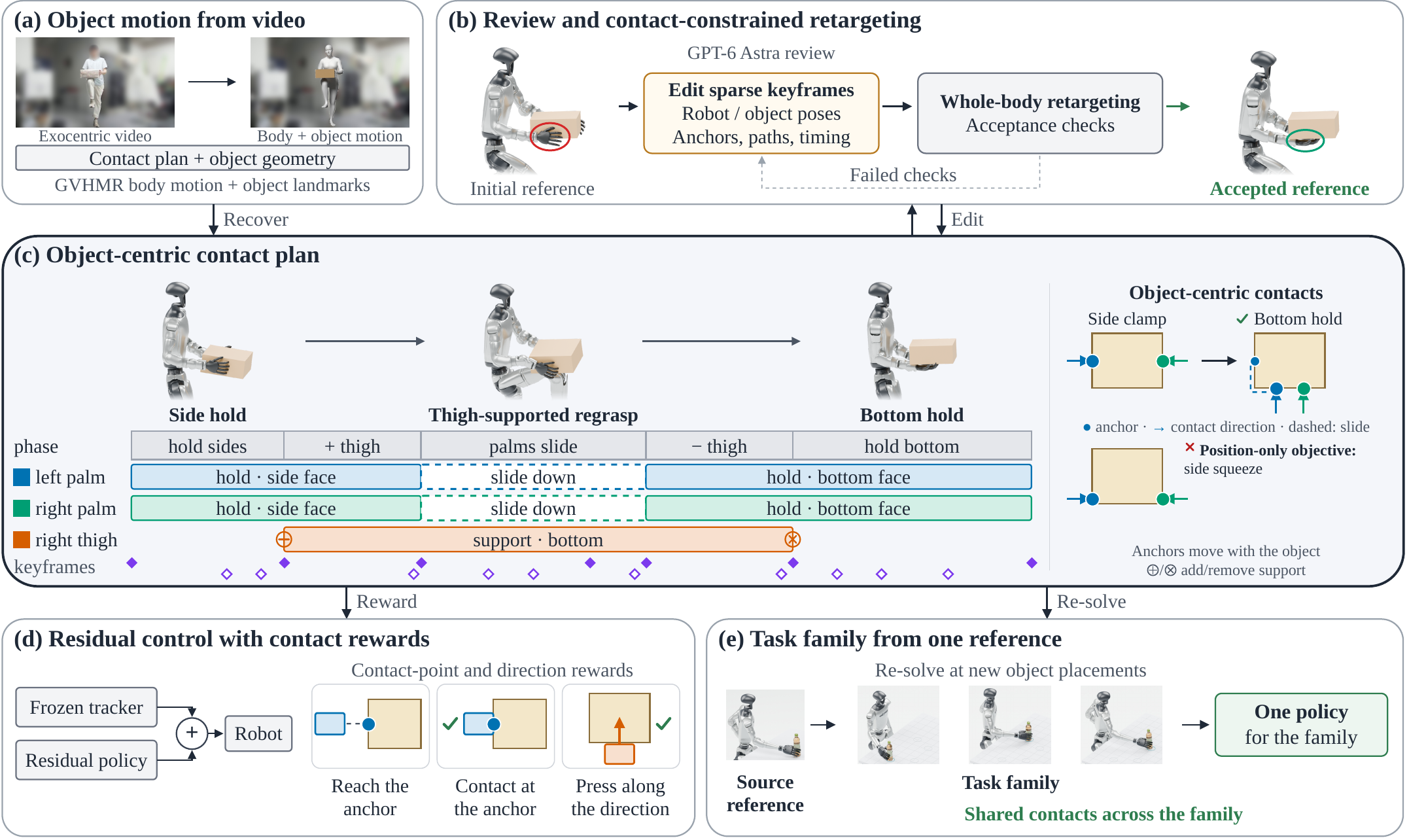}
\caption{Overview of ReContact. (a) An exocentric video supplies body
and object motion; Fig.~\ref{fig:egocentric} shows the egocentric route.
(b) GPT-6 Astra reviews sparse keyframes against the video, and
contact-constrained whole-body optimization turns them into
humanoid--object references. (c) The object-centric contact plan, here for the thigh-supported regrasp, lists, per phase, the
paired robot--object surfaces with their contact points, contact
directions and sticking or sliding modes; adding or removing a pair gives a contact transition. Keyframes of the run shown: contact keyframes where
support changes ($\blacklozenge$) and motion keyframes inside phases
($\lozenge$). An objective that does not specify the pressing
direction also accepts a squeeze near the lower side edges ($\times$). (d) The plan defines the rewards of a
residual controller for contact at the planned point and along the
planned direction. (e) Re-solving the placement-dependent contact segment and
its approach at new object placements gives a task family for one policy.}
\label{fig:framework}
\end{figure*}

\IEEEPARstart{H}{umanoid} manipulation often requires an object's
support to pass between the hands and other body surfaces, so that the
hands can reposition or become free for another action: a box held by
its sides, rested on a raised thigh while the palms slide underneath; 
a tube tucked under the arm to free one hand. Such a change of the contacts that
support an object is a \emph{contact transition}. Contact transitions
are difficult to transfer from human video to a humanoid: body occlusion
degrades the estimated object pose, the robot's kinematic structure and
dimensions differ from the human's, and a policy can reproduce the
reference object trajectory while contacting it at different points. Successful transfer therefore requires explicit contact locations,
pressing directions and the order of support changes.

Existing methods carry some of these elements.
PRISM, HOI-Retarget and OTRetarget preserve contact locations during
retargeting~\cite{prism,hoiretarget,otretarget};
ResMimic supervises whether each body link is in
contact~\cite{resmimic};
CHORD compares pressing directions through wrench
analysis~\cite{chord};
and WBD learns the order of support changes among predefined
holds~\cite{wbd}.
However, no method carries the locations, pressing directions and
support-change order together across reconstruction, retargeting and
control. The learned policy may therefore contact the object differently from the video. 
For example, a side squeeze and a bottom hold on a box place the palms at similar positions, but
 their pressing directions differ: sideways versus upward.

We propose ReContact, which reconstructs and retargets human contacts
through an object-centric \emph{contact plan} (Fig.~\ref{fig:framework}).
For each phase, the plan specifies the active robot--object contact
pairs. Each pair carries a contact anchor, a pressing direction and a contact mode.
 For exocentric video, GPT-6 Astra drafts the
plan; for egocentric video, the plan is defined by the hand--object
contacts reconstructed following AGILE~\cite{agile}.
The plan addresses the difficulties in turn:
contact priors from the plan tie the object pose to the supporting body
surfaces through occlusion; whole-body optimization adapts the posture
to the robot while keeping the planned anchors and directions; and a
residual policy over a frozen motion tracker~\cite{resmimic,sonic} is
rewarded for realized contact at each anchor and along its pressing
direction. The plan also marks the placement-dependent segment, which is
re-solved at new object placements and re-anchored to new object sizes
to turn one reference into a \emph{task family}, a set of references for one task, 
each trained by a single policy~\cite{grail}.

We evaluate ReContact on an
under-arm box carry, the thigh-supported regrasp, under-arm holds of a
tissue pack and a tube, and grasps from egocentric video. In
simulation, with one reference shared by all controllers, ReContact
completes the thigh-supported regrasp into the prescribed two-palm
bottom hold in $\TtwoSuccessMean\pm\TtwoSuccessSD\%$ of episodes over
three training seeds; the adapted baselines complete
$\TtwoComparatorCompletionMin$--$\TtwoComparatorCompletionMax\%$ of
episodes but form the hold in at most $\TtwoBestComparatorMean\%$.
Training one policy on a $247$-reference family raises tube-transfer
success inside the generated workspace from $20.4\%$, for a policy
trained on the source reference alone, to $97.8\%$. On a Unitree G1, 
the learned policies execute seven contact-rich manipulation tasks and two loco-manipulation tasks.
 Our contributions are:
\begin{itemize}
\item An object-centric contact plan that specifies, for each phase, the
contact pairs with their anchors, pressing
directions and sticking or sliding modes, and the support changes
between phases.
\item Contact-constrained object reconstruction and whole-body
retargeting from exocentric or egocentric video that preserve the
plan's support relations, anchors and directions, with references
re-solved at new object placements and sizes for task-family training.
\item A residual controller rewarded for realized contact by each
designated link near its planned anchor and along its planned pressing
direction.
\item Real-world validation on a Unitree G1 over nine tasks, including a 
four-skill loco-manipulation chain from table pickup to floor placement.
\end{itemize}

\section{Related Work}
\label{sec:related_work}

\subsection{Reconstructing Human--Object Interaction from Video}
Reconstructing manipulation requires consistent human and object motion
despite occlusion and changing contacts. CARI4D and RePHO recover
exocentric interactions through object tracking and contact or physics
refinement~\cite{cari4d,repho}, and SUGAR refines motion and per-part
contact labels with a privileged physics refiner~\cite{sugar}. GRAIL
generates the interaction with a video model, rendering a
robot-proportioned character inside a specified 3D scene~\cite{grail},
and GenHOI imposes object-centric contact constraints on generated
video~\cite{genhoi}. For hand-centered input, AGILE reconstructs
hand--object interactions from monocular video~\cite{agile}, the
procedure ReContact adopts for egocentric clips; Do as I Do transfers
reconstructed interactions to dexterous robot hands~\cite{doasido}, and
OKAMI adapts one RGB-D demonstration to new object
locations~\cite{okami}.

Contact also constrains the reconstruction itself. CARI4D penalizes
hand--object distance where contact is labeled, RePHO rewards hand
contact states during physics-based refinement, and PRISM couples object
motion to palm anchors that it also uses in retargeting and
control~\cite{prism}. HumanX synthesizes object motion from a predefined
body anchor~\cite{humanx}, and SUGAR labels the task-specific body parts
in contact with the object. These couplings are defined on hand
contacts, on one predefined anchor or on per-part contact labels. In
manipulation with successive body supports, the object may move with a
thigh or forearm before the hands take it again, and its relation to
each supporting surface changes as the palms slide and supports are
released. ReContact models these relations through the carriers and
contacts of each phase of its contact plan, so that sticking and sliding
constraints guide the object pose through each contact transition.

\subsection{Contact-Preserving Retargeting}
Contact-preserving retargeting extends motion transfer from matching
body poses to maintaining interactions with objects and the environment.
UMR transfers contact maps through learned point-cloud
correspondences~\cite{umr}; HOI-Retarget preserves object-frame palm
contacts and world-frame palm orientations over temporal windows and
lets a user correct contact segments estimated from monocular
video~\cite{hoiretarget}; and OTRetarget jointly adjusts robot and
object poses to preserve surface distances and
directions~\cite{otretarget}. Contact retargeting for extrinsic
manipulation composes contact constraints across manipulation
primitives for a robot arm~\cite{extrinsiccontact}, OmniRetarget
preserves body--object--terrain relations through an interaction
mesh~\cite{omniretarget}, and VICAR enforces manually annotated
world-frame contact poses~\cite{vicar}. These methods keep the
demonstrated interaction geometry; HOI-Retarget, for instance, keeps
each palm contact from slipping on the object. In support-changing
manipulation, the geometry must also stay consistent as contacts are
established, moved and released: transferring the thigh-supported
regrasp involves not only placing the thigh and palms against the box
but also coordinating the palm slide with the withdrawal of the thigh.
ReContact represents these changes as successive contact phases with
contact anchors or sliding paths, pressing directions and sticking or
sliding modes; retargeting adjusts the robot-side contact points and the
whole-body posture while keeping the planned progression of support.

\begin{table*}[t]
\centering
\caption{Comparison of Humanoid Manipulation Methods and How They
Specify Contacts and Whether They Generate References}
\label{tab:related}
\footnotesize
\setlength{\tabcolsep}{3.5pt}
\renewcommand{\arraystretch}{1.1}
\providecommand{\cyes}{\checkmark}
\providecommand{\cno}{--}
\begin{tabular}{@{}L{0.125\textwidth}L{0.072\textwidth}L{0.36\textwidth}cccc@{}}
\toprule
Method & Data & Contact representation
 & \shortstack{Body\\support} & Direction & \shortstack{Stick/\\slide}
 & \shortstack{Reference\\generation}\\
\midrule
HDMI~\cite{hdmi} & Video, MoCap
 & Hand and foot points in the object frame; per-frame contact flag
 & \cno & \cno & \cno & \cno\\
HumanX~\cite{humanx} & Video
 & Body-anchor--object transform; binary contact labels
 & \cno & \cno & \cno & \cyes\\
PRISM~\cite{prism} & Video
 & Palm anchors in the object frame, fixed within a phase
 & \cno & \cno & \cno & \cyes\\
VICAR~\cite{vicar} & Video
 & Annotated hand and foot poses in the world frame
 & \cno & \cno & \cno & \cyes\\
CHORD~\cite{chord} & Video, MoCap
 & Hand contact points and normals as an object-centric wrench space
 & \cno & \cyes & \cno & \cno\\
OmniRetarget~\cite{omniretarget} & MoCap
 & Interaction mesh over body joints, object and terrain points
 & \cno & \cno & \cno & \cyes\\
HOI-Retarget~\cite{hoiretarget} & Video, MoCap
 & Palm and foot targets in the object frame, with palm orientation
 & \cno & \cyes & \cno & \cno\\
OTRetarget~\cite{otretarget} & MoCap, RGB-D video
 & Signed distance, closest point and direction of every surface probe near another surface
 & \cno & \cyes & \cno & \cno\\
ResMimic~\cite{resmimic} & MoCap
 & Binary contact per torso, hip and arm link from link--object distance
 & \cyes & \cno & \cno & \cno\\
InterMimicGen~\cite{intermimicgen} & MoCap
 & Per-hand contact intent; link-to-object-surface vectors; thumb--finger opposition
 & \cno & \cno & \cno & \cyes\\
WBD~\cite{wbd} & Synthetic
 & Grasp mode (left, both or right arm) and object position on predefined
   body regions; surface normals in grasp synthesis
 & \cyes & \cyes & \cno & \cno\\
\midrule
\rowcolor{ourrow}\textbf{ReContact} & Video
 & Point and pressing direction of each designated link on an object anchor, per phase
 & \cyes & \cyes & \cyes & \cyes\\
\bottomrule
\end{tabular}
\par\smallskip
\begin{minipage}{\textwidth}\footnotesize
\emph{Data}: source of the references or training data; MoCap: motion
capture; Synthetic: holds optimized in simulation and policy rollouts,
without human demonstrations. \checkmark: reported in the cited work;
--: not reported. \emph{Body support}: a contact on a body surface other than the hands
and feet, such as a forearm, thigh or torso, is specified as an object
support. \emph{Direction}: each contact specifies the direction in which
the robot presses on the object, as a contact normal or force direction.
\emph{Stick/slide}: each contact is declared as sticking or as sliding
along the object surface. \emph{Reference
generation}: one demonstration is expanded into new references for
changed object placements or sizes, and a policy is trained on them.
\end{minipage}
\end{table*}

\subsection{Contact-Guided Whole-Body Control}
Whole-body manipulation has progressed from maintaining body-assisted
grasps to repositioning objects across body surfaces. TACT, Punyo and
embracing controllers use body contacts to handle large
objects~\cite{tact,punyoegrl,embracingbulky}. WBD synthesizes grasps on
predefined arm and torso regions from the object's surface normals and
learns transitions among left-arm, both-arm and right-arm holds,
including under-arm stows~\cite{wbd}; its goals specify an object
position and a grasp mode and let the object roll and slide across the
body between goals, which differs from reproducing a support sequence
recovered from video. WoCoCo commands sequential contact
goals~\cite{wococo}, and SceneBot, ContactMimic and OmniContact bring
per-link or binary contact states into whole-body
control~\cite{scenebot,contactmimic,omnicontact}.

In motion imitation, contact rewards add physical interaction to
kinematic tracking. ResMimic and ManipTrans refine pretrained motion
through residual control~\cite{resmimic,maniptrans}. PRISM and HDMI
combine contact locations with force magnitude~\cite{prism,hdmi}, HumanX
uses binary hand contact states~\cite{humanx}, ResMimic rewards
contact-force magnitude on the torso, hip and arm links labeled in the
reference, DemoMimic rewards hand-normal alignment and contact
persistence in dexterous manipulation~\cite{demomimic}, and CHORD
compares the wrench capability of hand contacts~\cite{chord}. These
objectives reward physically meaningful interaction, but a successful
object motion can still admit different contact configurations: in a
regrasp, a side squeeze near the lower edges can complete the object
trajectory without passing support to the palms underneath. ReContact
uses the contact plan to separate such executions, rewarding contact by
each designated link near its planned anchor and normal force along its
planned pressing direction.

\subsection{Reference Generation}
Reference generation extends demonstrations to new task configurations.
MimicGen transforms object-relative end-effector segments and keeps the
executions that succeed~\cite{mimicgen}; HumanoidMimicGen adds
whole-body skill-start planning and torso--arm
adaptation~\cite{humanoidmimicgen}, and DemoHLM transforms end-effector
trajectories, object-centric before manipulation and body-centric during
it~\cite{demohlm}. InterReal offsets the object and wrists and re-solves
the arms~\cite{interreal}, and VLK synthesizes interactions in
reconstructed scenes~\cite{vlk}. Other methods adapt the interaction
geometry directly: OmniRetarget, VICAR and HumanX re-solve a reference
for new objects, terrain or contact
targets~\cite{omniretarget,vicar,humanx}, HOI-Retarget scales the object
of one demonstration and OTRetarget substitutes
it~\cite{hoiretarget,otretarget}, PRISM expands its data through
counterfactual videos~\cite{prism}, and InterMimicGen edits the
placement and posture, fine-tunes its tracker on the edited references
and keeps those it executes~\cite{intermimicgen}. For a skill with body
supports, a new placement changes the reach to the object, while the
contacts that define the skill, on the hands and on the body, must stay
as demonstrated. ReContact reads this split from the contact plan: it
re-solves the placement-dependent segment at each new placement, rejoins
the source motion or a common hold pose, and trains one policy on the
resulting task family. Table~\ref{tab:related} compares what a contact
specifies in the closest methods and whether they generate references
from one demonstration; ReContact is the only method in the table that
specifies body supports, pressing directions and stick/slide modes and
also generates references.

\section{Method}
\label{sec:method}

Given a monocular video of a person manipulating a rigid object and a
target humanoid, ReContact builds a robot--object reference and learns a
policy that executes its contacts (Fig.~\ref{fig:framework}).

\subsection{Object-Centric Contact Plan}
\label{sec:program}
A contact plan describes how the robot supports and manipulates the
object over time, separately from its whole-body posture. Each phase
specifies the active contact pairs (each a robot surface region paired with an object surface region), their contact anchors
and pressing directions, their sticking or sliding modes and the surface
that carries the object; phase boundaries mark changes in these support
relations or contact modes. For exocentric video, GPT-6 Astra drafts
the phases and pairs, and an operator can review and correct them
in the same viewer (Section~\ref{sec:reconstruction}); for egocentric
video, the recovered hand--object segment forms the plan with the hand
as the carrier. Fig.~\ref{fig:framework}(c) shows one plan.

For $K$ phases, the plan is
\begin{equation}
\begin{aligned}
\mathcal P &= \{I_k,\mathcal C_k,\zeta_k,\mathcal H_k\}_{k=1}^{K},\\
c &= (\ell_c,\mathbf p_c^\ell,\mathbf m_c^\ell,
       \mathbf a_c(\tau),\mathbf n_c(\tau),\sigma_c).
\end{aligned}
\label{eq:contact_program}
\end{equation}
Here $I_k=[s_k,e_k]$ spans phase $k$ from its start frame $s_k$ to its
end frame $e_k$, $\tau=(t-s_k)/(e_k-s_k)$ is its normalized progress at
frame $t$, and $\mathcal C_k$ is its set of active pairs. The carrier
$\zeta_k$ names the scene surface or robot link, a hand included, on
which the object rests or with which it moves. The stable subinterval
$\mathcal H_k\subset I_k$ excludes contact closure and release; the
training-assistance gate (Section~\ref{sec:assist}) evaluates contact
quality there.

Each pair binds a point $\mathbf p_c^\ell$ and normal
$\mathbf m_c^\ell$ in the frame of robot link $\ell_c$ to a contact
anchor $\mathbf a_c$ and outward surface normal $\mathbf n_c$ in the
object frame. The contact direction $-\mathbf n_c$ is the direction
along which the robot presses on the object. The mode
$\sigma_c\in\{\text{stick},\text{slide}\}$ keeps the anchor fixed or
moves it along the surface path $\mathbf a_c(\tau)$. Posture keyframes
and foot-support intervals accompany the plan, and each contact records
its visual evidence, a confidence and whether it was observed or
inferred under occlusion.

The fields are filled in stages. The video stage places each anchor on
a named object face in fractions of the object's half extents; object
fitting uses the plan's supports, carriers and anchor faces with the
human body, and the object dimensions convert the face coordinates into
metric anchors. Retargeting then sets the robot-side points
$\mathbf p_c^\ell$ and normals $\mathbf m_c^\ell$ and adapts the posture
to the robot. The object-side anchors, paths and normals remain the
contact targets through retargeting and policy training.

\subsection{Object Motion and Contact Plan from Video}
\label{sec:reconstruction}
Both video routes recover object geometry and motion, scene supports
and a contact plan. Exocentric input also supplies body motion and
posture keyframes; for egocentric input, retargeting generates the body
motion around the hand--object interaction (Section~\ref{sec:retarget}).

\subsubsection{Contact Plan from Exocentric Video}
Gravity-view human motion recovery (GVHMR)~\cite{gvhmr} estimates
world-frame body motion.
Object dimensions come from the video or from a measured deployment
object. The object, the ground plane and support surfaces such as a
table are then placed in the world frame of the estimated body motion. GPT-6 Astra drafts the phase boundaries, the object's
rest intervals on scene supports and the foot-swing intervals, and adds
approach, turn and posture keyframes. For each pair it names the human body part, which fixes the robot link $\ell_c$; the stick/slide mode; and the object anchor as a point on a named face, in fractions of the
object's half extents, with that face's outward normal; a sliding pair
receives a path of such points. It also marks visible object landmarks,
which short bidirectional optical flow propagates for rigid-object
fitting.

\subsubsection{Object Pose from Contact Priors}
\label{sec:object_pose}
To bridge occlusion, ReContact couples object motion to the body through
three contact priors, extending PRISM's palm constraints~\cite{prism}.
\emph{Support} keeps a resting object on its scene surface.
\emph{Carrier} keeps its pose constant relative to a hand, thigh or
forearm carrier:
\begin{equation}
\widehat{\mathbf T}_o(t)
  =\widehat{\mathbf T}_{\mathrm{car}}(t)
     \boldsymbol\Delta_{\mathrm{car},o}.
\label{eq:phase_repair}
\end{equation}
The world poses $\widehat{\mathbf T}_o$ and
$\widehat{\mathbf T}_{\mathrm{car}}$ determine a relative transform
$\boldsymbol\Delta_{\mathrm{car},o}$ estimated on visible frames and
propagated through occlusion. \emph{Slide} fixes contact anchors at the
phase's endpoints and interpolates between them under the remaining
visual evidence. Landmarks constrain visible spans; the priors bridge
occluded spans. One least-squares fit combines landmark reprojection,
temporal smoothness, a prior on the measured object size and these
priors. We construct these priors from the plan’s supports, carriers, 
and anchor faces, together with the estimated human body motion. 
On egocentric video the hand--object segment is recovered
following AGILE~\cite{agile}, with the hand as the carrier.

\subsubsection{Review}
Reconstruction and retargeting alternate only through review. Within a
round the object track is fitted first and then held fixed while the
robot is fitted to it, so the dependency runs one way. GPT-6 Astra then
reads the fitted object drawn over the source frames with its largest
landmark residuals, the robot and object rendered from the video camera
over the same frames, and the failed acceptance checks with their frames
(Section~\ref{sec:retarget}). It corrects the object pose through its
evidence, by re-placing or adding landmarks, marking occluded ones as
inferred, and revising rest intervals, supports, anchors or contact
intervals; it also edits phase boundaries and keyframes. The next round
refits the track and the robot from the revised plan. An operator can
inspect and correct every element in the same viewer.

\subsection{Contact-Constrained Whole-Body Retargeting}
\label{sec:retarget}
Retargeting preserves the contact plan while adapting posture and
free-space motion. For exocentric input, UMR~\cite{umr} initializes the
posture from GVHMR, and confidence-weighted human limb directions and
torso orientation serve as posture targets. Egocentric input initializes
the hands from retargeted hand poses. Each named robot link receives a
surface point and normal facing its object anchor; the point may move on
the link to suit the robot's geometry. The face-fraction anchors become
metric at the fitted object size.

\subsubsection{Surface-Constrained Whole-Body Fitting}
The fit optimizes the floating pelvis, waist, legs, arms and available finger joints. 
Object poses are not optimized: an object on a scene surface keeps its reviewed pose, 
and a carried object moves rigidly with its fitted carrier link.
 For object pose $\mathbf T_o^\star(t)$ and robot-link pose
$\mathbf T_\ell(\mathbf q_t)$, contact residuals are
\begin{equation}
\begin{aligned}
\mathbf e^p_{ct}
 &=\mathbf T_{\ell_c}(\mathbf q_t)\mathbf p_c^\ell
   -\mathbf T_o^\star(t)\mathbf a_c(\tau),\\
\mathbf e^n_{ct}
 &=\mathbf R_{\ell_c}(\mathbf q_t)\mathbf m_c^\ell
   +\mathbf R_o^\star(t)\mathbf n_c(\tau).
\end{aligned}
\label{eq:anchor_pair}
\end{equation}
The residuals bring the points together and oppose the normals, leaving rotation about the normal free for posture adaptation.

The retargeting objective groups terms by function:
\begin{equation}
\begin{aligned}
\mathcal E_{\mathrm{ret}}
 &=E_{\mathrm{contact}}+E_{\mathrm{posture}}
   +E_{\mathrm{stance}}\\
 &\quad+E_{\mathrm{smooth}}+E_{\mathrm{collision}}.
\end{aligned}
\label{eq:reference_compilation}
\end{equation}
Contact terms penalize \eqref{eq:anchor_pair}; posture terms keep the
human limb directions and keyframe targets; stance terms fix supporting
feet; temporal terms coordinate neighboring states; collision penalties
cover robot, object and scene. Joint limits and finger couplings are enforced.
Static whole-body solves first fit the keyframes: one at every contact
boundary and the motion keyframes that GPT-6 Astra places where the
motion changes inside a phase (Fig.~\ref{fig:framework}(c)). Each returns in seconds and identifies binding
contacts when the requested configuration is unreachable.

\subsubsection{Planning around Egocentric Interaction Segments}
\begin{figure*}[t]
\centering
\includegraphics[width=\textwidth]{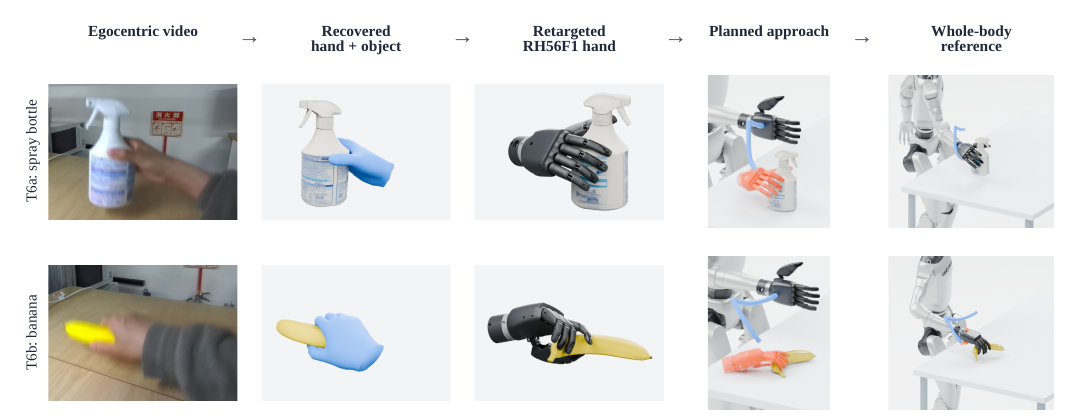}
\caption{Egocentric route, shown for a spray bottle and a banana.
Recovered hand--object segments are retargeted to the Inspire RH56F1
hand; planning generates the approach and the remaining G1 body motion.
Blue paths mark the approach, orange hands the grasp targets.
Recovered views use the source camera and crop; the last column shows
the composed reference at the same source frame.}
\label{fig:egocentric}
\end{figure*}

Collision-checked rapidly-exploring random tree (RRT-Connect)
paths~\cite{rrtconnect} with minimum-jerk
timing connect the
robot's start pose to an egocentric interaction segment and back
(Fig.~\ref{fig:egocentric}).

\subsubsection{From Keyframes to the Dense Reference}
Interpolating the keyframes in robot joint space, and the object and
anchors in phase progress, initializes a dense fit. Acceptance checks on
every frame and on intermediate states test contact alignment, palm
orientation, penetration, joint limits, speed and stance; overlays and
limb-direction errors test agreement with the source. A failed check
names the keyframe or surface assignment to revise. The accepted
reference is
\begin{equation}
\mathcal R=
 (\mathbf q^\star,\dot{\mathbf q}^\star,
  \mathbf T_o^\star,\mathcal P),
\label{eq:compiled_reference}
\end{equation}
so control uses the same contact pairs that shaped the motion.

\subsection{Residual Learning with Contact-Point and Direction Rewards}
\label{sec:tracker}
A frozen motion tracker supplies whole-body targets, and a residual
policy corrects their execution to establish the planned contacts.

\subsubsection{Residual Action and Observations}
Let $\mathbf q_t^{\rm B}$ be the frozen tracker's body targets and
$\mathbf q_{h,t}^{\star}$ the reference hand coordinates. The policy
produces scaled corrections $\Delta\mathbf q_{b,t}$ and
$\Delta\mathbf q_{h,t}$, giving
\begin{equation}
\begin{aligned}
\mathbf q^{\rm cmd}_{b,t}
 &=\operatorname{clip}_{\mathcal Q_b}
   (\mathbf q_t^{\rm B}+\Delta\mathbf q_{b,t}),\\
\mathbf q^{\rm cmd}_{h,t}
 &=\operatorname{clip}_{\mathcal Q_h}
   (\mathbf q_{h,t}^{\star}+\Delta\mathbf q_{h,t}).
\end{aligned}
\label{eq:residual_action_short}
\end{equation}
Here $b$ and $h$ denote body and hand coordinates, and
$\mathcal Q_b,\mathcal Q_h$ are their admissible command ranges.
Commands are filtered and rate-limited before application.
The actor receives proprioception, action history, robot/object
references, the phase and the reference's contact mask, which marks the
pairs scheduled at the current frame, kinematic hand feedback, and the
task's object-pose stream. The critic additionally receives
privileged dynamics and contact information. Policies are trained with proximal policy optimization (PPO)~\cite{ppo}.

\subsubsection{Contact-Point and Contact-Direction Reward}
For an active pair $c$, let $d_c$ be the distance between the robot
contact point and the object anchor under the current robot and object
poses. A physical contact matches the plan when the simulator reports
contact between link $\ell_c$ and the object within a matching radius of
the anchor; $g_c\in\{0,1\}$ indicates such a matched contact. Let
$\mathbf f_c$ be the sum, in object coordinates, of the normal forces
that the matched contacts exert on the object. The pair reward is
\begin{equation}
\begin{aligned}
m_c
 &=\exp\!\left[-\beta_n
   \left(1-(-\mathbf n_c)^\top
          \frac{\mathbf f_c}{\|\mathbf f_c\|}\right)\right],\\
r_c
 &=\alpha_c e^{-\beta_c d_c}
   +g_c m_c e^{-\beta_c d_c}+\eta_c\rho_c.
\end{aligned}
\label{eq:contact_reward_short}
\end{equation}
We set $m_c=0$ when $\|\mathbf f_c\|=0$. The first term rewards
proximity to the anchor, with weight $\alpha_c$ and distance decay
$\beta_c$. The second rewards an actual contact at the anchor whose force
points along the contact direction $-\mathbf n_c$, with angular decay
$\beta_n$. The third rewards persistence: $\rho_c=1$ when the pair is
matched at both the current and the preceding control step, with weight
$\eta_c$.

Pair rewards are aggregated with task contact weights and combined with
object, body, keypoint and finger tracking and with regularization of
residuals and applied targets.

\begin{algorithm}[t]
\caption{Task-family generation from one reference}
\label{alg:family}
\begin{algorithmic}[1]
\Require source reference $\mathcal R_0$, contact plan $\mathcal P$,
  start pose $\mathbf q_s$, support region $\mathcal W$, size $N$
\Ensure task family $\mathcal F$
\State Read from $\mathcal P$ the placement-dependent segment
  $[s_c,e_m)$: first hand-pair closure to the first body carrier or hold
\State Tile $\mathcal W$ into $N$ equal-area cells with centroids
  $\mathbf g_i$; $\mathcal F\gets\emptyset$
\For{$i=1,\dots,N$}
  \For{each object-symmetry and approach candidate, smallest change first}
    \State Place the object at $\mathbf g_i$; move the anchors and normals of $\mathcal P$ with it
    \State Re-solve waist, arm, pelvis and legs on $[s_c,e_m)$: contact
      targets, fixed feet, center of mass in support, limits, clearance
    \State Fade the placement offset to zero at $e_m$, or converge to a common hold pose
    \State Connect $\mathbf q_s$ to frame $s_c$ by a joint-space blend or a planned path
    \If{the kinematic and static-balance audit passes}
      \State $\mathcal F\gets\mathcal F\cup\{\mathcal R^{(i)}\}$; \textbf{break}
    \EndIf
  \EndFor
\EndFor
\end{algorithmic}
\end{algorithm}

\subsubsection{Contact-Aware Training Assistance}
\label{sec:assist}
A training-only virtual-object controller, as in
DexMachina~\cite{dexmachina} and ResMimic~\cite{resmimic}, applies a
proportional--derivative wrench that pulls the object toward its
reference, so that early policies learn to hold it. A gated decay
curriculum withdraws the wrench, lowering its gain when object tracking,
contact adherence and normal error over the stable subintervals
$\mathcal H_k$ and the reference completion rate pass their thresholds.
All evaluations run without assistance.

\begin{figure*}[!t]
\centering
\input{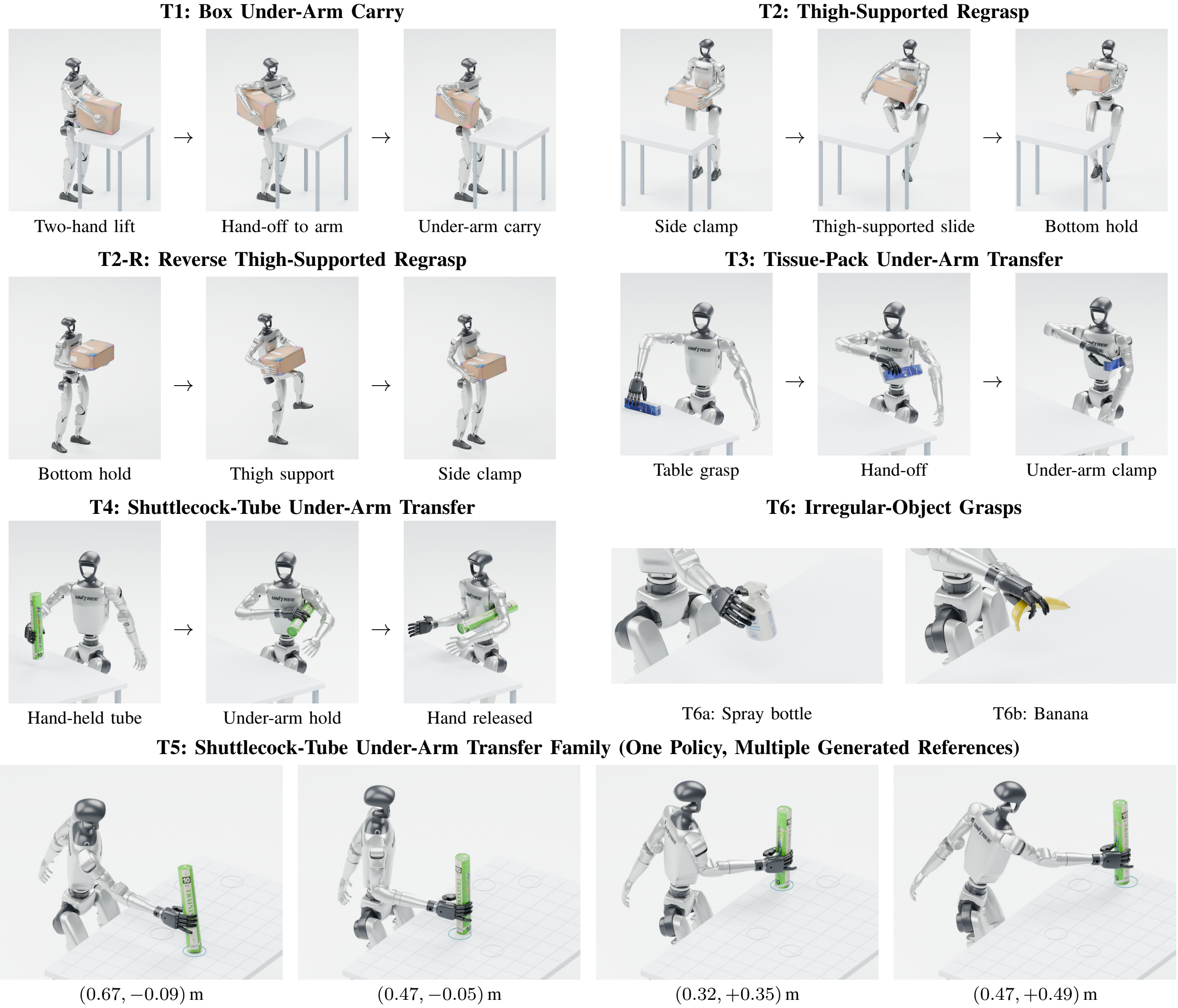}
\caption{Evaluated tasks in simulation; arrows give the temporal order.
T5 shows one policy trained on $247$ generated references grasping the
tube at four of their positions (tube $xy$ in the world frame).}
\label{fig:tasks}
\end{figure*}
\subsection{Task Families and Skill Composition}
\label{sec:datagen}
Because the anchors live on the object, the plan also specifies how
a reference adapts when the object placement changes
(Algorithm~\ref{alg:family}). Its phases delimit the segment that
depends on the placement: the segment begins when the first hand pair
closes. It ends when the carrier first becomes a body link, 
as the torso does in an under-arm hold, or at the first steady hold. 
At each declared
placement, the object-side anchors and normals move with the object, and
a whole-body optimization re-solves the waist, the acting arm, the pelvis
and the legs over this segment, with the recorded feet fixed, the center
of mass inside the support polygon and the posture pulled toward the
source. The placement offset fades out before body-carried phases, which
keep the source motion, or the members converge to a common hold pose;
a joint-space blend or a planned path connects the shared start pose to
the re-solved segment. Every member passes a kinematic and static-balance
audit of contact error, penetration, clearance, joint limits and
smoothness. This produces a task family
$\{\mathcal R^{(i)}\}_{i=1}^{N}$ before policy training. One residual
policy is trained across the family, each reference paired with its
object placement, so that both the manipulation posture and the approach
follow the object and the residual only corrects local execution errors.

The same mechanism adapts to object size. For box-shaped objects,
re-anchoring preserves face identities and normalized in-face anchor
coordinates, so the anchors move with the faces of the resized object;
general meshes use explicit surface correspondences.

For composition with walking, family reference $i$ stores the object
pose relative to its reference pelvis, $\mathbf T_{po}^{(i)}$.
Given a live object pose $\mathbf T_o$, the pelvis station of reference $i$,
the pelvis pose from which its manipulation starts, is
\begin{equation}
\mathbf T_p^{(i)}
 =\mathbf T_o\bigl(\mathbf T_{po}^{(i)}\bigr)^{-1}.
\label{eq:reference_station}
\end{equation}
Pose-based selection chooses the reference with the smallest normalized
planar position and yaw error. The locomotion controller approaches
the station, the manipulation reference is aligned to the live pelvis,
and the residual is blended in while the tracker state is retained.

\begin{table*}[t]
\centering\footnotesize
\caption{Shared-Reference Controller Comparison in Simulation}
\label{tab:success}
\setlength{\tabcolsep}{5pt}
\textbf{(a) T2: thigh-supported regrasp}\par\smallskip
{\scriptsize\setlength{\tabcolsep}{3pt}
\begin{tabular}{lrrrrrrrr}
\toprule
Method & Complete & SR & Thigh & Bottom L & Bottom R & Normal$\downarrow$ & $E_o\downarrow$ & Mech. work$\downarrow$\\
 & (\%) & (\%)$\uparrow$ & (\%) & (\%) & (\%) & (deg) & (cm) & (J)\\
\midrule
HDMI-style & $24.3\pm20.0$ & $0.08\pm0.07$ & $0.2\pm0.3$ & $10.6\pm16.1$ & $3.5\pm4.2$ & $47.8\pm4.6$ & $11.24\pm1.06$ & $565.2\pm90.1$\\
ResMimic-style & $65.7\pm5.8$ & $0.01\pm0.02$ & $20.6\pm5.6$ & $0.3\pm0.4$ & $4.5\pm7.1$ & $55.0\pm3.0$ & $9.47\pm1.29$ & $868.0\pm21.1$\\
CWS adapter & $52.3\pm9.2$ & $0.16\pm0.24$ & $36.6\pm25.1$ & $0.6\pm0.5$ & $65.7\pm56.2$ & $49.2\pm2.8$ & $9.25\pm2.08$ & $787.1\pm59.9$\\
\rowcolor{ourrow}Ours & $\mathbf{89.0\pm8.6}$ & $\mathbf{83.52\pm5.20}$ & $\mathbf{94.4\pm2.8}$ & $\mathbf{80.8\pm1.8}$ & $\mathbf{81.2\pm7.6}$ & $\mathbf{16.3\pm4.5}$ & $\mathbf{5.54\pm0.64}$ & $\mathbf{544.9\pm20.2}$\\
\bottomrule\end{tabular}}\par\medskip
\textbf{(b) T1, T3, T4, T6a and T6b: under-arm carry, under-arm holds and grasps}\par\smallskip
\begin{tabular}{llrrrrrr}
\toprule
Task & Method & SR (\%)$\uparrow$ & $E_o$ (cm)$\downarrow$ & Anchor (cm)$\downarrow$ & Adherence (\%)$\uparrow$ & Normal (deg)$\downarrow$ & Mech. work (J)$\downarrow$\\
\midrule
T1 & HDMI-style & $92.9$ & $8.96\pm1.17$ & $7.53\pm1.34$ & $6.0\pm3.3$ & $15.6\pm5.3$ & $1169\pm29$\\
 & ResMimic-style & $100.0$ & $7.13\pm1.12$ & $9.22\pm1.21$ & $3.8\pm2.1$ & $51.8\pm1.3$ & $1419\pm33$\\
 & CWS adapter & $100.0$ & $8.88\pm0.86$ & $12.78\pm1.31$ & $\mathbf{12.6\pm9.5}$ & $\mathbf{1.1\pm0.7}$ & $\mathbf{1328\pm32}$\\
\rowcolor{ourrow} & Ours & $100.0$ & $\mathbf{5.89\pm0.79}$ & $\mathbf{7.32\pm1.23}$ & $11.4\pm9.0$ & $3.7\pm3.3$ & $1354\pm29$\\
\cmidrule(l){1-8}
T3 & HDMI-style & $97.8$ & $2.93\pm0.66$ & $4.15\pm1.21$ & $28.5\pm14.8$ & $20.1\pm6.2$ & $118\pm2$\\
 & ResMimic-style & $99.9$ & $\mathbf{1.39\pm0.20}$ & $3.42\pm0.31$ & $44.0\pm6.2$ & $24.0\pm3.3$ & $228\pm5$\\
 & CWS adapter & $100.0$ & $1.49\pm0.14$ & $\mathbf{2.28\pm0.10}$ & $40.0\pm2.6$ & $17.8\pm1.2$ & $154\pm1$\\
\rowcolor{ourrow} & Ours & $100.0$ & $1.87\pm0.21$ & $2.29\pm0.25$ & $\mathbf{44.5\pm1.5}$ & $\mathbf{15.9\pm1.1}$ & $\mathbf{153\pm2}$\\
\cmidrule(l){1-8}
T4 & HDMI-style & $97.3$ & $3.22\pm0.72$ & $2.85\pm2.08$ & $61.7\pm20.0$ & $15.1\pm4.2$ & $287\pm8$\\
 & ResMimic-style & $96.8$ & $11.24\pm0.58$ & $4.27\pm0.47$ & $52.5\pm4.2$ & $16.6\pm2.0$ & $514\pm21$\\
 & CWS adapter & $100.0$ & $1.15\pm0.16$ & $1.77\pm0.22$ & $56.3\pm7.0$ & $\mathbf{8.8\pm1.5}$ & $\mathbf{202\pm4}$\\
\rowcolor{ourrow} & Ours & $100.0$ & $\mathbf{0.87\pm0.10}$ & $\mathbf{1.03\pm0.11}$ & $\mathbf{83.4\pm0.7}$ & $11.8\pm1.2$ & $211\pm4$\\
\cmidrule(l){1-8}
T6a & HDMI-style & $99.6$ & $2.85\pm0.22$ & $4.53\pm0.10$ & $13.4\pm1.3$ & $11.9\pm2.2$ & $160\pm3$\\
 & ResMimic-style & $99.9$ & $0.74\pm0.08$ & $3.06\pm0.09$ & $46.9\pm3.1$ & $7.5\pm0.3$ & $\mathbf{117\pm3}$\\
 & CWS adapter & $99.8$ & $0.80\pm0.09$ & $3.60\pm0.09$ & $31.8\pm1.0$ & $\mathbf{5.0\pm1.0}$ & $127\pm3$\\
\rowcolor{ourrow} & Ours & $99.9$ & $\mathbf{0.70\pm0.09}$ & $\mathbf{2.91\pm0.07}$ & $\mathbf{53.0\pm3.5}$ & $6.5\pm1.2$ & $126\pm3$\\
\cmidrule(l){1-8}
T6b & HDMI-style & $99.7$ & $2.18\pm0.19$ & $1.91\pm0.14$ & $51.9\pm6.5$ & $44.1\pm5.9$ & $139\pm3$\\
 & ResMimic-style & $99.9$ & $1.31\pm0.17$ & $3.05\pm0.26$ & $34.6\pm5.8$ & $26.4\pm6.7$ & $176\pm4$\\
 & CWS adapter & $99.9$ & $\mathbf{1.00\pm0.07}$ & $1.85\pm0.07$ & $48.0\pm5.3$ & $\mathbf{24.7\pm4.4}$ & $\mathbf{125\pm3}$\\
\rowcolor{ourrow} & Ours & $99.9$ & $1.05\pm0.08$ & $\mathbf{1.57\pm0.08}$ & $\mathbf{86.9\pm4.0}$ & $33.4\pm1.3$ & $130\pm3$\\
\bottomrule\end{tabular}
\end{table*}

\subsection{Implementation Details}
\label{sec:method_impl}
\subsubsection{Visual Initialization and Acceptance Checks}
The egocentric front end follows AGILE~\cite{agile} with Segment
Anything Model~2 (SAM~2) masks~\cite{sam2}, MoGe-2 depth~\cite{moge2}, Hunyuan3D object
geometry~\cite{hunyuan3d}, WiLoR estimates of the parametric MANO hand
model~\cite{wilor,mano},
and FoundationPose~\cite{foundationpose} object-pose candidates.
Acceptance checks use an
explicit geometry-pair list covering self-collision and contact with the
object, supports and ground, including pairs disabled in the
simulator's native collision masks.

\subsubsection{Tracker and Learning Configuration}
SONIC~\cite{sonic} is adapted once from its released $29$-joint layout
to the robot's $23$ body joints using Any2Any's joint mapping and
low-rank fine-tuning~\cite{any2any}, and then frozen. The residual
policy controls the $23$ body joints and, when the robot carries an
articulated hand, its active finger joints.

\section{Experiments}
\label{sec:experiments}
The experiments answer four research questions (RQs). RQ1: Does the
ReContact controller, rewarded for contact at the planned point and
along the planned direction, execute the planned contact transition rather than
any hold that keeps the object on track? RQ2: Do reconstruction and
retargeting under the plan keep the video's contacts while the whole
body adapts to the robot and to a new object size? RQ3: Does expanding one reference into a task family extend
the workspace beyond start randomization? RQ4: Do the learned skills run
on the physical robot and compose with locomotion?

\subsection{Experimental Setup}
\label{sec:t3_protocol}

The tasks differ in contact sequence, object geometry, supporting body
surfaces and video viewpoint; each instantiates the contact plan of
Section~\ref{sec:program} with its own phases and contact pairs
(Fig.~\ref{fig:tasks}). T1--T3 come from exocentric video and T6 from
egocentric video. T1 carries a box under the arm. T2 moves an already
side-clamped box over the thigh to a two-palm bottom hold; T2-long adds
table pickup and retreat, and T2-R reverses the transfer. T3 moves a
tissue pack from the hand to an under-arm clamp. T4 replaces the tissue pack of 
T3 with a shuttlecock tube and keeps the same
contact plan, from the hand grasp to the under-arm hold, in which the
left upper arm and torso close on the tube.
T5 expands the T4 reference into a family of $247$ references over
tube placements on the table (Section~\ref{sec:exp_datagen}). T6a and T6b
grasp a spray bottle and a banana without a body support. On T3--T6 the
robot carries the right Inspire RH56F1 hand, whose six active joints the
residual policy also controls; T1/T2 use passive rubber hands without
finger joints.

Table~\ref{tab:success} compares ReContact with three baselines
adapted to the same reference for each task
(Section~\ref{sec:exp_contact}). All controllers simulate the G1 of
Section~\ref{sec:method_impl} in mjlab, and all compared policies train
for $3000$ updates of $4096\times24$ transitions. Each task is evaluated
with the object randomization reached in training; T5 is evaluated over
the position grid of Section~\ref{sec:exp_datagen}.
Table~\ref{tab:success}(a) uses three independently trained policies per
method, with seeds $42$--$44$ and three batches of $1000$ evaluation
episodes per policy; entries report the mean and sample standard
deviation across the three policy means. Within each policy, the thigh,
bottom-contact, normal-error and mechanical-work columns average completed
episodes, and $E_o$ averages all finite per-episode measurements,
including failures. Table~\ref{tab:success}(b) uses one frozen policy
per row and $3000$ evaluation episodes, and its $\pm$ values are
standard deviations across episodes; error and adherence metrics use all
finite per-episode measurements, including failures, and mechanical work
uses successful episodes.

\begin{figure*}[tp]
\centering
\includegraphics[width=\textwidth]{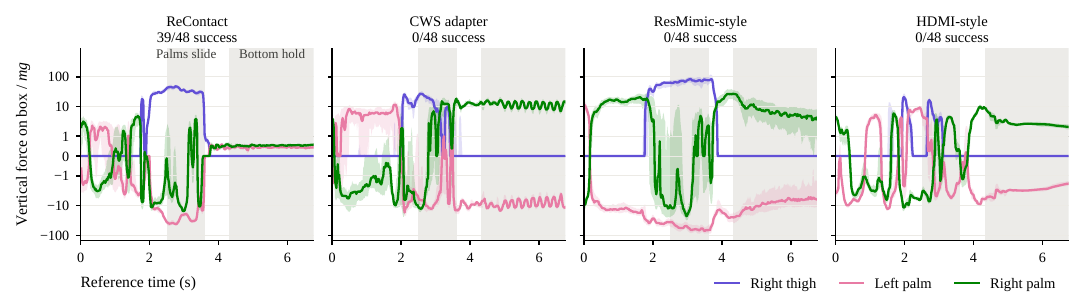}
\caption{Only ReContact moves the T2 box from thigh support to a two-palm
bottom hold. Vertical contact force (normal plus friction) exerted on the
box by the right thigh and each palm, divided by the box weight $mg$
(dashed line), in $48$ episodes per controller. Lines are medians and bands interquartile ranges over the episodes
alive at each reference time ($0.1$~s moving median within each
episode); the axis is logarithmic beyond $\pm mg$. Shading marks the palm
slide and the terminal bottom hold. Panel titles give two-palm successes.}
\label{fig:t2_support}
\end{figure*}
\begin{figure*}[tp]
\centering
\includegraphics[width=\textwidth]{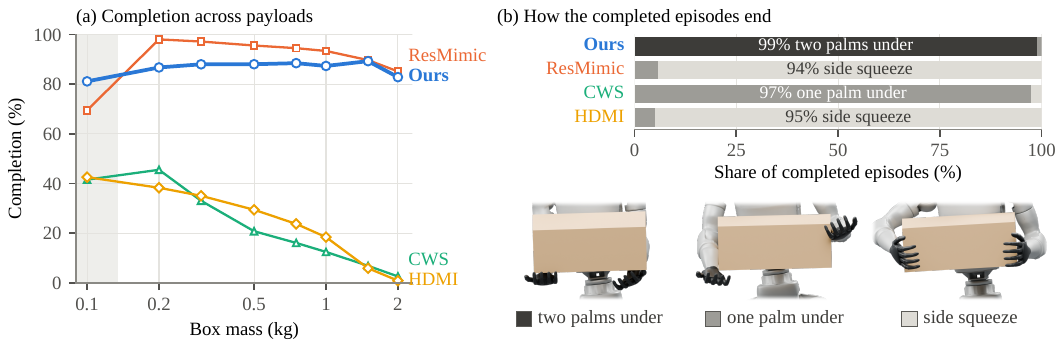}
\caption{The planned contact transition persists across payload variation.
(a) Completion of ReContact and the ResMimic-style, HDMI-style and CWS controllers at eight payloads, $3000$ episodes per method and mass; shading marks the
mass randomization used in training ($0.074$--$0.135$~kg around the $0.1$~kg nominal mass). (b) Completed episodes pooled over the payloads, split by the bottom-hold criterion: met by both palms (the T2 success predicate), by one palm, or by neither (side squeeze). Images: terminal holds of completed $0.5$~kg episodes (ReContact, CWS, ResMimic-style).}
\label{fig:t2_payload}
\end{figure*}
\begin{figure*}[!t]
\centering
\includegraphics[width=\textwidth]{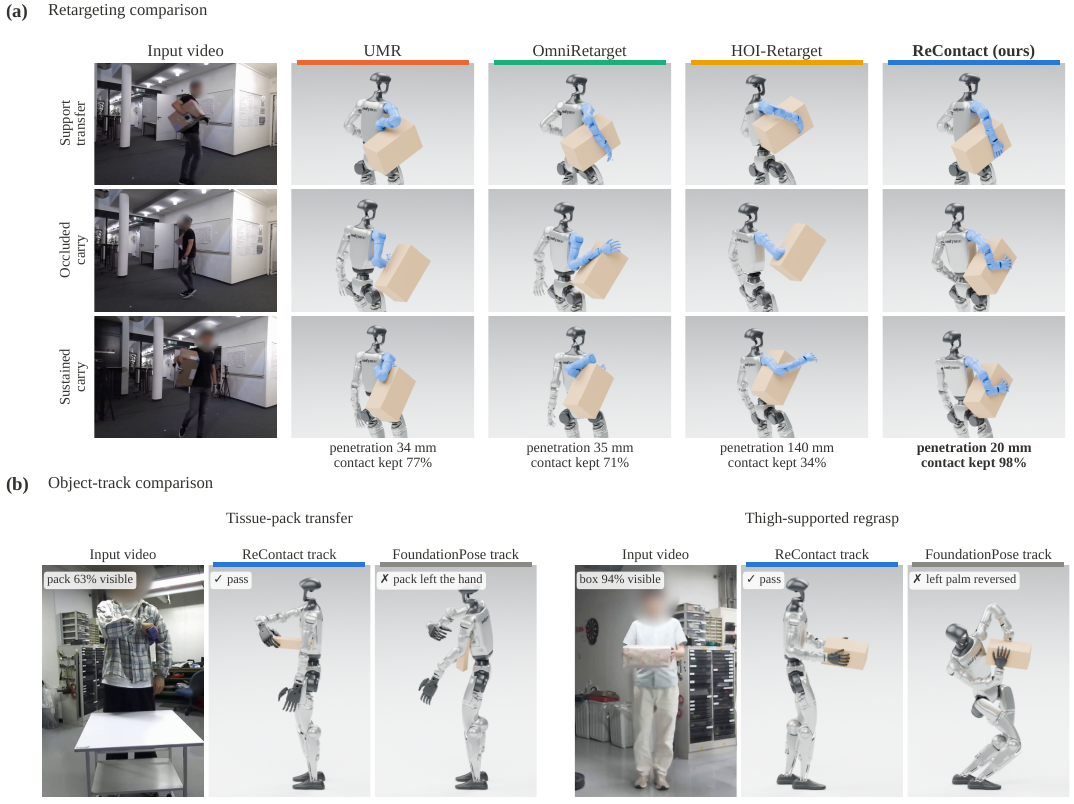}
\caption{Contact preservation in retargeted humanoid--object references.
(a)~Comparison of retargeting methods on a medium-box carry using shared
body motion and object tracks. (b)~Comparison of object tracks for the
tissue-pack transfer and thigh-supported regrasp using the same ReContact
retargeting procedure.}
\label{fig:shared_evidence}
\end{figure*}

Complete gives the fraction of episodes that finish without a failure
termination; success further requires the task's contact outcome, and
the success rate (SR) is the fraction of successful episodes. T1
requires completion of the carry. T2 requires simultaneous bottom-face
contact of both planned palms for at least half of the terminal hold
stage. T3 and T4 end with the object held under the left arm after the
right hand releases; success requires the left arm and torso to clamp
the object, with a mean opposing normal load of at least $1$~N, and
right-hand load below $0.25$~N over the last $0.5$~s, so an object retained under the arm counts as
successful despite an offset from the reference pose. T5 adds a
placement-accuracy requirement to the T4 predicate
(Section~\ref{sec:exp_datagen}); T6 requires an $8$~cm lift during its
manipulation phase. The same task predicate is used for every method.

Beyond SR, Table~\ref{tab:success} reports object tracking, contact
fidelity and mechanical work. $E_o$ averages world-frame error over
$1000$ object-surface points. Anchor error includes missed contacts.
Adherence is the fraction of scheduled link--frame samples realized as
normal-force contact within $30$~mm of the anchor; on T2, Thigh applies
it to the planned thigh contact, and Bottom L and Bottom R give the
fraction of the terminal hold stage in which the left or right palm
contacts the bottom face. Normal error measures the angle between a
matched contact force and its planned direction $-\mathbf n_c$.
Mechanical work integrates absolute actuator power. Bold indicates the
best value per task; mechanical work is compared only among methods with
the task's highest success rate.

\subsection{Realizing the Planned Contact Transition (RQ1)}
\label{sec:exp_contact}
\label{sec:exp_tracker}

\subsubsection{Shared-Reference Controllers}
The baselines differ from ReContact only in the controller. ResMimic-style~\cite{resmimic} uses the frozen SONIC
residual architecture with a coarse contact-force objective and an
assistance schedule. HDMI-style~\cite{hdmi} uses a reference-conditioned
specialist with random-phase training without assistance and its
method-specific losses and terminations. The
contact-wrench-space (CWS) adapter inserts CHORD's
per-hand, object-centric wrench-space objective~\cite{chord} into the
ReContact residual setup and assistance schedule and adds missed- and
unintended-contact penalties.

Table~\ref{tab:success}(a) separates completing the motion from executing
its contact transition. Across training seeds, the baselines' mean completion
rates range from $\TtwoComparatorCompletionMin$ to $\TtwoComparatorCompletionMax\%$,
but the best baseline reaches $\TtwoBestComparatorMean\%$ two-palm
success, compared with ReContact's $\TtwoSuccessMean\pm\TtwoSuccessSD\%$.
The per-palm fractions show
how the baselines end: in completed CWS episodes the right palm is on
the bottom face for $\TtwoCwsBottomRight\%$ of the hold and the left for $\TtwoCwsBottomLeft\%$,
whereas ReContact keeps both palms there ($\TtwoOursBottomLeft/\TtwoOursBottomRight\%$). Its thigh adherence is $\TtwoOursThigh\%$, against $\TtwoComparatorThighMin$--$\TtwoComparatorThighMax\%$ for the baselines. Its contact
forces deviate from the planned directions by $\TtwoOursNormal^\circ$, against
$\TtwoComparatorNormalMin$--$\TtwoComparatorNormalMax^\circ$, and it has the lowest mean
all-episode $E_o$. Among completed episodes it also
has the lowest mean mechanical work ($\TtwoOursWork$ against $\TtwoComparatorWorkMin$--$\TtwoComparatorWorkMax$~J).
Fig.~\ref{fig:t2_support} follows the box's vertical support through
the contact transition in $48$ episodes per controller.
While the palms slide, the ReContact thigh pushes the box upward with a
median $35.0\,mg$ against the left palm's downward $26.8\,mg$, clamping it on
the thigh; once the thigh releases, both palms push it upward ($0.44$ and
$0.53\,mg$), the planned two-palm bottom hold. ResMimic-style also loads the
thigh during the slide, but no baseline reaches that hold: in their bottom
holds the left palm pushes down while the right pushes up, squeezing the box
between the hands ($-9.0/12.1$, $-9.0/7.0$ and $-2.9/2.6\,mg$ for CWS,
ResMimic-style and HDMI-style). For the baselines the side squeeze is
a local optimum: leaving it risks the box and staying in it is still
rewarded. Sliding a palm to the bottom loosens the side grip, and losing
the box ends the episode, the cause of $84$--$99\%$ of each baseline's
early stops. A side palm still earns their contact rewards:
ResMimic-style credits force anywhere on the object, HDMI-style scores
force near targets beside the lower side edges, and CWS keeps
$0.27$--$0.29$ of its bottom-hold value. ReContact's reward gives a side
contact less than $4\%$ of the planned bottom contact, so staying pays
almost nothing and ReContact forms the two-palm bottom hold in all three
training seeds.

Table~\ref{tab:success}(b) compares contact fidelity on tasks that most
controllers complete. On T4, ReContact places the tube closest to its
anchors ($1.03$~cm against at least $1.77$~cm) and has the highest adherence ($83.4\%$ against at most $61.7\%$). ResMimic-style
also retains the tube under the left arm, with a larger reference-pose
offset. On the irregular objects of T6a and T6b,
ReContact also has the lowest anchor error and the highest adherence.
On T3 its anchor error matches that of CWS ($2.29/2.28$~cm), and its
contact forces deviate least from the planned directions ($15.9^\circ$).
The CWS adapter gives the lowest normal error on T1, T4, T6a and T6b
($8.8^\circ$ against ReContact's $11.8^\circ$ on T4). The two objectives
weight direction differently: the CWS objective scores each hand's
realized contact wrench against the planned one directly, whereas the
ReContact direction term counts only for a contact matched within the
radius of its anchor and decays with the distance to it. Consistent with this difference, the
ReContact policies reach higher adherence on T4, T6a and T6b, while the CWS policies align the hand forces more closely.

T1 tests the carry rather than contact-point placement. ReContact
completes $2999$ of $3000$ carries and keeps the box closest to its
reference ($E_o=5.89$~cm against at least $7.13$~cm), yet its adherence is $11.4\%$. The
scheduled links do touch the box: the hands in $96\%$ of their scheduled
samples and the right forearm and torso in $83\%$; within $30$~mm of the
anchor these fractions fall to $18\%$ and $5\%$. The low adherence
therefore comes from contacts made elsewhere on the box, not from missed
contacts. Across episodes, anchor error rises with object error (Pearson
$r=0.80$), so the $7.32$~cm anchor error, the lowest of the four
controllers, follows the displacement of the carried box. With every
controller at $3.8$--$12.6\%$ adherence, T1 compares carry completion
and object tracking; contact placement is compared on the other tasks.

The residual policy is required to execute the planned contacts. We
evaluate the frozen SONIC tracker with a zero residual action, as in
ResMimic's tracker-only baseline~\cite{resmimic}. Its success rate is
zero on every task of Table~\ref{tab:success}.

\subsubsection{Contact Transition across Payloads}
\label{sec:t2_payload}
The payload sweep tests whether the contact transition holds for loads far
outside training. The four policies share the T2 reference and a
$0.36\times0.20\times0.14$~m box whose mass is randomized in training
between $0.074$ and $0.135$~kg around its $0.1$~kg nominal value; the
sweep fixes the mass at eight values from $0.1$ to $2.0$~kg. ReContact completes $81$--$89\%$ of episodes with no trend in mass,
whereas the CWS adapter and HDMI-style fall from $42$--$43\%$ at
$0.1$~kg to below $3\%$ at $2.0$~kg (Fig.~\ref{fig:t2_payload}(a)).
ResMimic-style completes the most episodes, $70$--$98\%$, but completion
does not say how the box is held. Under the T2 success predicate,
completion with both palms under the box, ReContact keeps
$78.9$--$88.2\%$ at every payload and the other controllers reach at most
$0.2\%$. Fig.~\ref{fig:t2_payload}(b) shows how their completions end:
ResMimic-style completes $90.4\%$ of the $24{,}000$ episodes yet establishes the two-palm hold in $6$ of them, and $94\%$ of its completions end without a palm meeting the bottom-hold criterion ($95\%$ for HDMI-style); the CWS adapter completes $22.4\%$, $97\%$ of them with one palm meeting it. ReContact establishes the two-palm hold
in $99\%$ of its completed episodes ($85.5\%$ of all episodes).

\subsection{From Video to Contact-Preserving References (RQ2)}
\label{sec:exp_reference}

\subsubsection{Contacts from Shared Inputs}
Fig.~\ref{fig:shared_evidence}(a) compares contact geometry during a
medium-box carry, a clip separate from the T1 reference video. Every method receives the same GVHMR body motion, object geometry, scene
and ReContact object track, and uses the 29-degree-of-freedom G1
model. Contact targets are fixed before the comparison:
each retargeter receives them in its native representation.
HOI-Retarget~\cite{hoiretarget} receives per-hand phase labels,
UMR's~\cite{umr} contact-map variant receives the human--object contact
map, and OmniRetarget~\cite{omniretarget} preserves its interaction
mesh. ReContact additionally uses its contact plan and posture
keyframes, with model-reviewed repair passes on its reference.
The views show the difference: the
retargeters place the arm over or through the box, whereas ReContact
cradles it on the forearm with the palm on the planned face. Measured on
the whole clip with one contact schedule for every method, the
retargeters penetrate the box by up to $34$--$140$~mm and keep the
scheduled arm within $30$~mm of the box in $34$--$77\%$ of the scheduled
frames; ReContact penetrates at most $20$~mm and keeps $98\%$.

\subsubsection{Object Track}
Fig.~\ref{fig:shared_evidence}(b) compares ReContact's object pose
estimation with FoundationPose~\cite{foundationpose} under the same
retargeting. In the tissue-pack transfer, the arm hides up to $72\%$ of the
pack. FoundationPose loses the pack's orientation and flips it by more
than $100^\circ$, so the retargeted
hand loses the pack. In the thigh-supported regrasp, the box is nearly symmetric, 
and FoundationPose's pose selection alternates between two depths $12$~cm apart 
along the camera ray, so the retargeted left palm turns over. ReContact ties the
object to the holding body part and keeps its pose stable in both clips,
and both retargeted references pass the acceptance checks.

\begin{figure}[!t]
\centering
\includegraphics[width=\columnwidth]{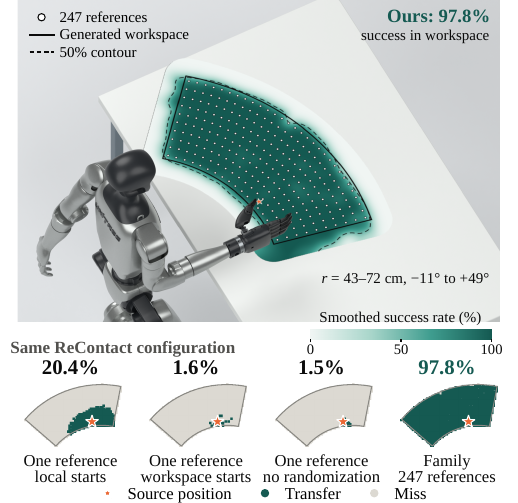}
\caption{Workspace coverage from one reference (T5 tube-transfer family). Top: success of the task-family policy, smoothed with a $1.5$~cm Gaussian kernel, inside the generated workspace (outline) and in the $6$~cm band outside it; white points mark the $247$ references and the star the source reference's tube position. Bottom: per-position outcomes of ReContact policies of the same configuration trained on the source reference (local, $\pm2.5$~cm; workspace-wide; no start randomization) and on the family.}
\label{fig:t3_sampling}
\end{figure}
\begin{table}[t]
\centering
\caption{T5 Tube-Transfer Success (\%) of Single-Reference Controls
and the Family Policy by Distance (cm) From the Source Tube Position}
\label{tab:family}
\footnotesize
\setlength{\tabcolsep}{3pt}
\begin{tabular}{llcccccc}
\toprule
References & Start & $0$--$10$ & $10$--$20$ & $20$--$30$ & $>30$ & All & Ext.\\
\midrule
Source & None & $11.8$ & $0.0$ & $0.0$ & $0.0$ & $1.5$ & $0.1$\\
Source & $\pm2.5$~cm & $95.5$ & $25.3$ & $0.1$ & $0.0$ & $20.4$ & $8.4$\\
Source & Workspace & $12.2$ & $0.0$ & $0.0$ & $0.0$ & $1.6$ & $0.3$\\
\rowcolor{ourrow}Family ($247$) & $\pm2.5$~cm & $\mathbf{99.3}$ & $\mathbf{98.0}$ & $\mathbf{96.6}$ & $\mathbf{98.6}$ & $\mathbf{97.8}$ & $\mathbf{19.5}$\\
\bottomrule
\end{tabular}
\end{table}
\begin{figure*}[t]
\centering
\input{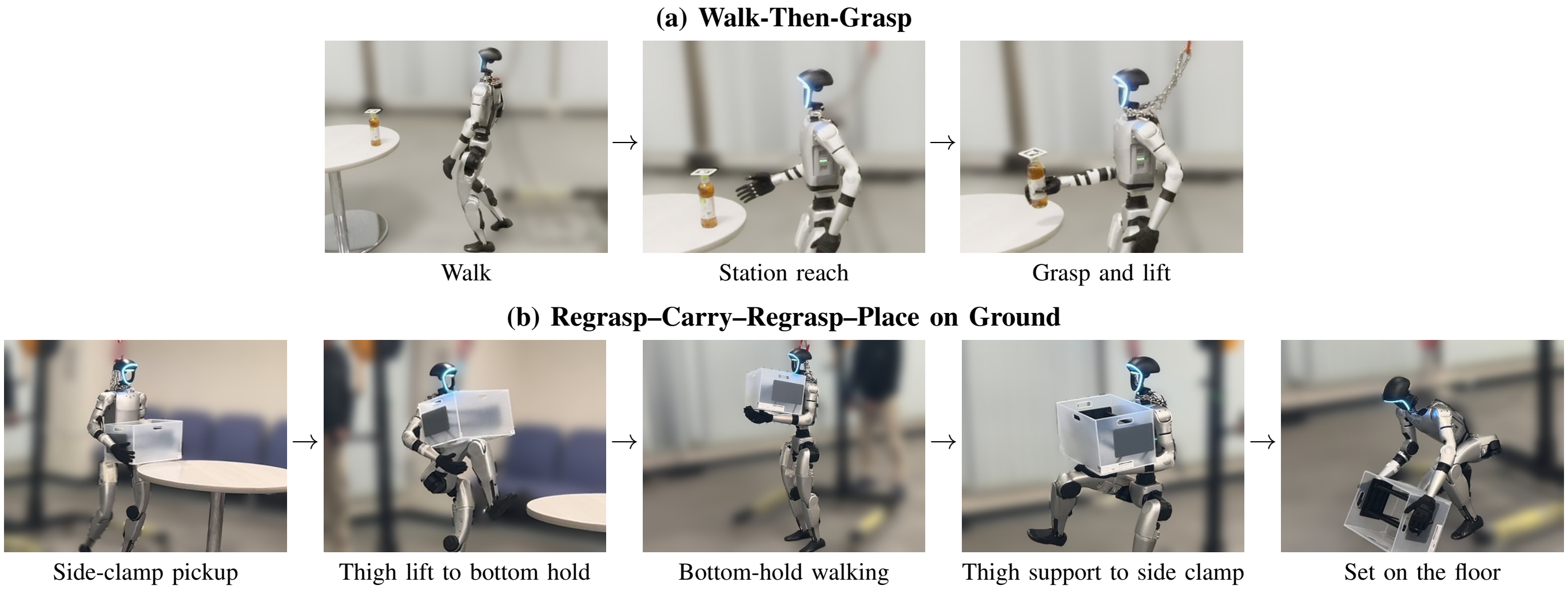}
\caption{Hardware deployment of learned contact skills on a Unitree G1.
(a) Walk-then-grasp using locomotion and the bottle-family policy.
(b) Table-to-floor box transfer combining forward and reverse
thigh-supported regrasps with loaded walking.}
\label{fig:composition}
\end{figure*}
\begin{figure*}[t]
\centering
\input{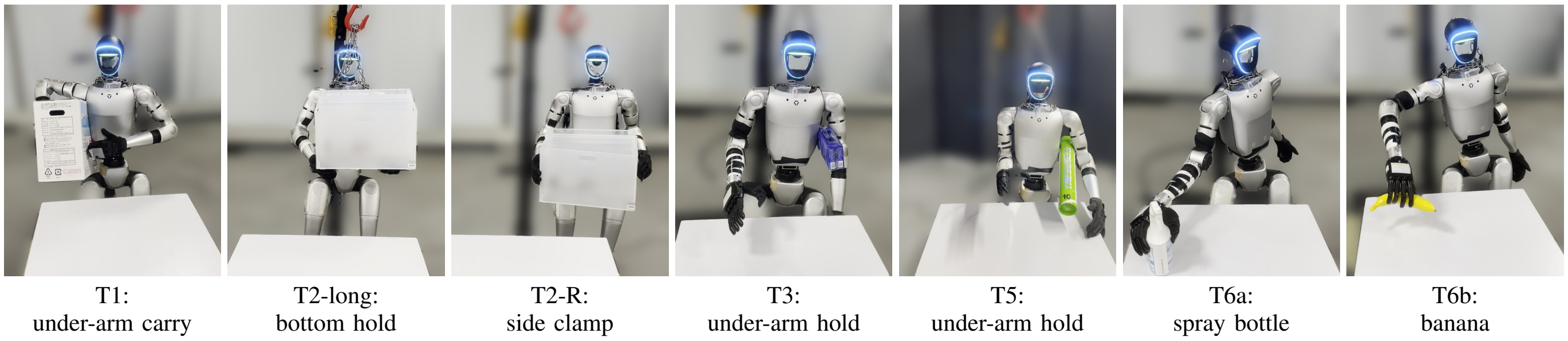}
\caption{Final hardware states of seven contact-rich manipulation tasks
on a Unitree G1.}
\label{fig:hardware}
\end{figure*}
\subsubsection{Retargeting to a Different Object Size}
The thigh-supported regrasp tests whether a plan carries over to an
object of another size. The T2 reference is built for box~A
($0.380\times0.270\times0.160$~m); re-anchoring its plan to the faces
of box~B ($0.389\times0.266\times0.236$~m, $7.6$~cm taller) gives the
box-B reference (Section~\ref{sec:datagen}), with the thigh support, the
palm slide and the two-palm bottom hold unchanged. A forward and reverse
policy pair is trained for each box, and both boxes complete the forward
and reverse regrasps on the G1 (Section~\ref{sec:real}).

\subsection{Expanding the Workspace from One Reference (RQ3)}
\label{sec:exp_datagen}
\label{sec:t3_sampling}

With the ReContact controller and its training configuration fixed,
expanding the T4 reference into the $247$ whole-body references of T5 raises tube-transfer success inside the generated workspace from $20.4\%$ for the best single-reference control to $97.8\%$
(Table~\ref{tab:family}).
Here, success additionally requires mean torso-frame object-position
error below $5$~cm over the final $0.5$~s to test placement accuracy.

The source reference places the tube $0.46$~m ahead of the pelvis and $0.07$~m to the left; the family tiles an annular sector of the table ($0.43$--$0.72$~m, $-11^\circ$ to $+49^\circ$, $1746$~cm$^2$) with $247$ equal-area cells, each solved for the whole body (Section~\ref{sec:datagen}). Training samples family references with $\pm2.5$~cm $xy$ offsets. Coverage is measured at $1743$ positions on a $1$~cm grid inside the workspace, including interpolation between references. Table~\ref{tab:family} groups them into four distance bins of $229$, $539$, $632$ and $343$ positions from the source reference's tube position and adds $1063$ positions in the $6$~cm exterior band (Ext.); each position is evaluated in three episodes with random tube mass and friction.

The single-reference controls train on the source reference with the same configuration and budget and differ only in start randomization (Start in Table~\ref{tab:family}): local starts reach $20.4\%$, workspace-wide starts $1.6\%$, and the Table~\ref{tab:success} T4 policy without start randomization $1.5\%$.
Generating references therefore extends coverage beyond randomizing
object starts around a fixed robot motion: with one reference, success stays concentrated around the source position (Fig.~\ref{fig:t3_sampling}), whereas the family policy
sustains $96.6$--$99.3\%$ in every distance bin. In the $6$~cm band
outside the generated workspace, success drops to $19.5\%$ for the
family policy and to at most $8.4\%$ for the controls
(Table~\ref{tab:family}), so the robot must walk to a pelvis station that puts the tube
inside the workspace.

\subsection{Hardware Demonstrations and Skill Composition (RQ4)}
\label{sec:exp_loco}
\label{sec:real}
We deploy the learned contact skills on a Unitree G1 and combine them
with locomotion (Fig.~\ref{fig:composition}). In walk-then-grasp
(Fig.~\ref{fig:composition}(a)), the robot approaches a table, stops at
the reference pelvis station, and uses the bottle-family policy to grasp
and lift the bottle. The longer sequence in Fig.~\ref{fig:composition}(b)
combines table pickup and the forward thigh-supported regrasp (T2-long),
a right turn and approximately $3$~m of loaded walking, the reverse
regrasp (T2-R), and floor placement. SONIC handles locomotion along a
precomputed route, while a separate residual policy performs the final
placement. Fig.~\ref{fig:hardware} shows the final states of seven
contact-rich manipulation tasks, including the two-palm bottom hold after
T2-long and the restored side clamp after T2-R. Additional hardware
demonstrations with boxes of different sizes and weights are provided in
the accompanying video.

The G1 runs the policies at $50$~Hz. T1 and walk-then-grasp use fiducial
markers~\cite{aruco} for object localization. T5 estimates the tube
position from the calibrated head RGB camera and selects the nearest
reference from its trained family. To simplify hardware deployment and
focus on contact execution rather than online object tracking, the
regrasp sequence uses nominal object poses latched at policy
initialization.

\section{Conclusion}
From monocular human video, ReContact extracts an object-centric contact plan 
that specifies which body surfaces support the object, 
in which directions they press on it, and how this support changes across phases. 
These specifications distinguish a
prescribed contact transition from other contacts that still retain the object.
In simulation, on a thigh-supported regrasp reference shared by all
controllers, the adapted baselines complete the motion while squeezing
the box between the palms, with at most $\TtwoBestComparatorMean\%$
two-palm success. This side squeeze is a reward local optimum: leaving
it risks the box, and staying in it is still rewarded. The ReContact
controller establishes the prescribed two-palm bottom hold after thigh
support and palm sliding in
$\TtwoSuccessMean\pm\TtwoSuccessSD\%$ of episodes over three training
seeds.

Anchoring contacts to the object also lets the placement-dependent
motion be re-solved at new object positions while the contact sequence
is preserved. Training one residual policy on the resulting reference
family raises tube-transfer success inside the generated workspace from
$20.4\%$ to $97.8\%$ relative to training on the single reference. On a
Unitree G1, the learned policies execute seven contact-rich manipulation
tasks and two loco-manipulation tasks, including a chain of table
pickup, the thigh-supported regrasp, a loaded walk of about $3$~m, the
reverse regrasp, and floor placement.

The system takes object geometry and pose inputs, a pretrained motion
tracker and one residual policy per task family. Future work will
investigate concurrent skills on different body surfaces, grasping while
walking and foot--terrain contacts, and a plan-conditioned controller
that executes contact transitions from new videos without retraining a policy.

\microtypesetup{activate=true}
\IEEEtriggeratref{41}
\bibliographystyle{IEEEtran}
\bibliography{references}

\clearpage
\end{document}